\documentclass[11pt,a4paper]{article}
\usepackage[hyperref]{acl2021}
\usepackage{times}
\usepackage{latexsym}
\usepackage{graphicx}
\usepackage{caption}
\usepackage{float}
\usepackage{amsmath}
\usepackage{multirow}
\usepackage{subfigure}

\usepackage{microtype}

\aclfinalcopy 

\title{CogAlign: Learning to Align Textual Neural Representations to Cognitive Language Processing Signals}

\author{
 Yuqi Ren and Deyi Xiong \Thanks{~Corresponding author} \\
 College of Intelligence and Computing, Tianjin University, Tianjin, China\\
 \texttt{ \{ryq20, dyxiong\}@tju.edu.cn} \\
 }
\date{}

\begin{document}
\maketitle
\begin{abstract}
Most previous studies integrate cognitive language processing signals (e.g., eye-tracking or EEG data) into neural models of natural language processing (NLP) just by directly concatenating word embeddings with cognitive features, ignoring the gap between the two modalities (i.e., textual vs. cognitive) and noise in cognitive features. In this paper, we propose a CogAlign approach to these issues, which learns to align textual neural representations to cognitive features. In CogAlign, we use a shared encoder equipped with a modality discriminator to alternatively encode textual and cognitive inputs to capture their differences and commonalities. Additionally, a text-aware attention mechanism is proposed to detect task-related information and to avoid using noise in cognitive features. Experimental results on three NLP tasks, namely named entity recognition, sentiment analysis and relation extraction, show that CogAlign achieves significant improvements with multiple cognitive features over state-of-the-art models on public datasets. Moreover, our model is able to transfer cognitive information to other datasets that do not have any cognitive processing signals. The source code for CogAlign is available at \url{https://github.com/tjunlp-lab/CogAlign.git}. 
\end{abstract}
\section{Introduction}

Cognitive neuroscience, from a perspective of language processing, studies the biological and cognitive processes and aspects that underlie the mental language processing procedures in human brains while natural language processing (NLP) teaches machines to read, analyze, translate and generate human language sequences \citep{DBLP:journals/corr/abs-2006-05113}. The commonality of language processing shared by these two areas forms the base of cognitively-inspired NLP, which uses cognitive language processing signals generated by human brains to enhance or probe neural models in solving a variety of NLP tasks, such as sentiment analysis \citep{mishra2017leveraging,barrett2018sequence}, named entity recognition (NER) \citep{DBLP:conf/naacl/HollensteinZ19}, dependency parsing \citep{strzyz2019towards}, relation extraction \citep{hollenstein2019advancing}, etc.

\begin{figure*}[ht] 
\centering
\includegraphics[width=.9\textwidth,height=7cm]{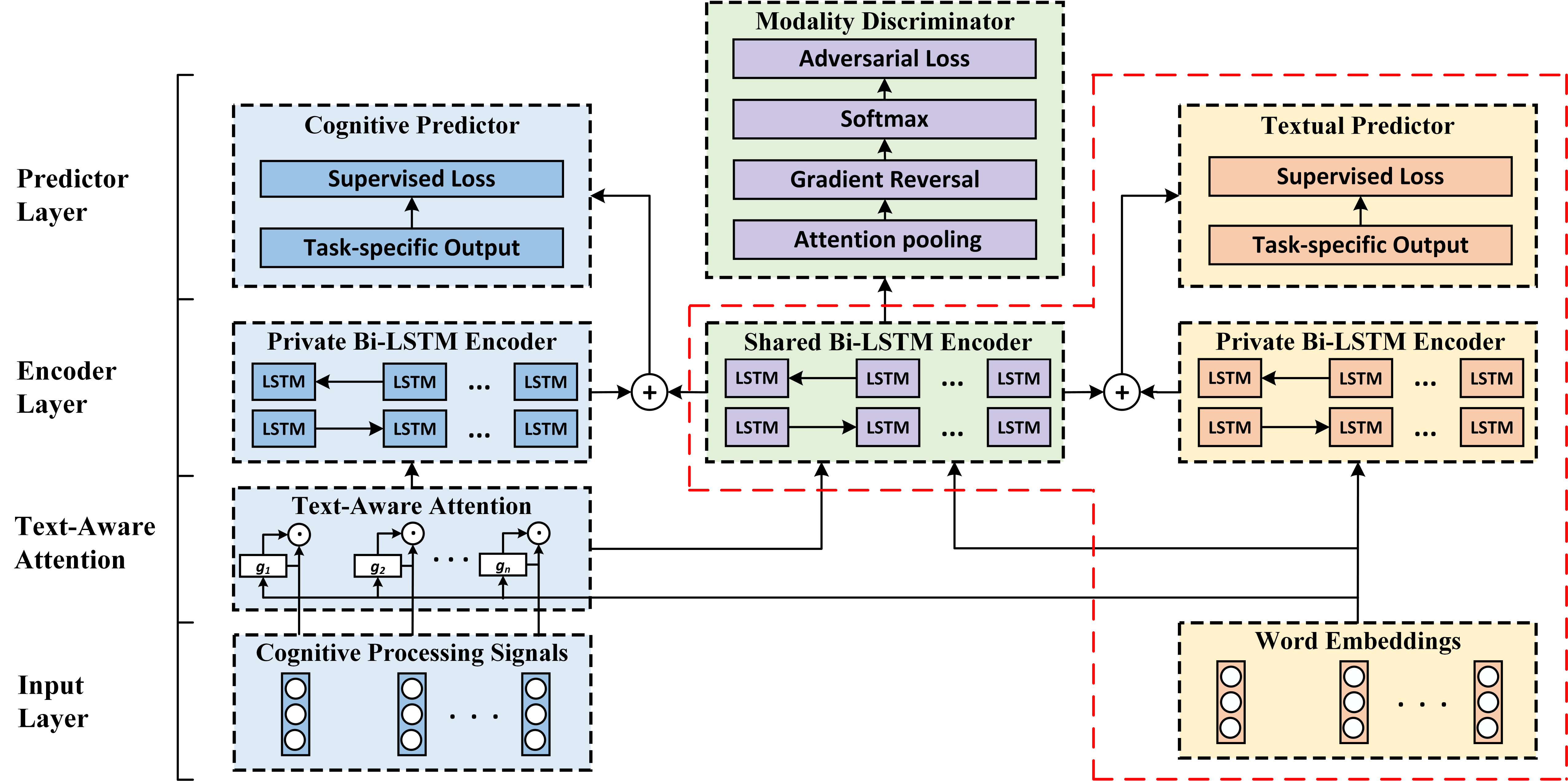}
\caption{Neural Architecture of the proposed CogAlign. For inference, only the components in the red dashed box are used.}\label{fig:basemodel}
\end{figure*}

In spite of the success of cognitively-inspired NLP in some tasks, there are some issues in the use of cognitive features in NLP. First, for the integration of cognitive processing signals into neural models of NLP tasks, most previous studies have just directly concatenated word embeddings with cognitive features from eye-tracking or EEG, ignoring the huge differences between these two types of representations. Word embeddings are usually learned as static or contextualized representations of words in large-scale spoken or written texts generated by humans. In contrast, cognitive language processing signals are collected by specific medical equipments, which record the activity of human brains during the cognitive process of language processing. These cognitive processing signals are usually assumed to represent psycholinguistic information \citep{2020A} or cognitive load \citep{antonenko2010using}. Intuitively, information in these two types of features (i.e., word embeddings and cognitive features) is not directly comparable to each other. As a result, directly concatenating them could be not optimal for neural models to solve NLP tasks. 

The second issue with the incorporation of cognitive processing signals into neural models of NLP is that not all information in cognitive processing signals is useful for NLP. The recorded signals contain information covering a wide variety of cognitive processes, particularly for EEG \citep{DBLP:journals/neuroimage/WilliamsKHWK19,DBLP:conf/sigir/EugsterRSKBRJK14}. For different tasks, we may need to detect elements in the recorded signals, which are closely related to specific NLP tasks, and neglect features that are noisy to the tasks. 

In order to address the two issues, we propose \textbf{CogAlign}, a multi-task neural network that learns to align neural representations of texts to cognitive processing signals, for several NLP tasks. As shown in Figure \ref{fig:basemodel}, instead of simply concatenating cognitive features with word embeddings, we use two private encoders to separately encode cognitive processing signals and word embeddings. The two encoders will learn task-specific representations for cognitive and textual inputs in two disentangled spaces. To align the representations of neural network with cognitive processing signals, we further introduce an additional encoder that is shared by both data sources. We alternatively feed cognitive and textual inputs into the shared encoder and force it to minimize an adversarial loss of the discriminator stacked over the shared encoder. The discriminator is task-agnostic so that it can focus on learning both differences and deep commonalities between neural representations of cognitive and textual features in the shared encoder. We want the shared encoder to be able to transfer knowledge of cognitive language processing signals to other datasets even if cognitive processing signals are not available for those datasets. Therefore, CogAlign does not require cognitive processing signals as inputs during inference. 

Partially inspired by the attentive pooling network \citep{santos2016attentive}, we propose a text-aware attention mechanism to further align textual inputs and cognitive processing signals at the word level. The attention network learns a compatibility matrix of textual inputs to cognitive processing signals. The learned text-aware representations of cognitive processing signals also help the model to detect task-related information and to avoid using other noisy information contained in cognitive processing signals. 

In a nutshell, our contributions are listed as follows:

\begin{itemize}
\item We present CogAlign that learns to align neural representations of natural language to cognitive processing signals at both word and sentence level. Our analyses show that it can learn task-related specific cognitive processing signals.
\item We propose a text-aware attention mechanism that extracts useful cognitive information via a compatibility matrix. 
\item With the adversarially trained shared encoder, CogAlign is capable of transferring cognitive knowledge into other datasets for the same task, where no recorded cognitive processing signals are available.
\item We conduct experiments on incorporating eye-tracking and EEG signals into 3 different NLP tasks: NER, sentiment analysis and relation extraction, which show CogAlign achieves new state-of-the-art results and significant improvements over strong baselines.
\end{itemize}

\section{Related Work}
\paragraph{Eye-tracking for NLP.}
Eye-tracking data have proved to be associated with language comprehension activity in human brains by numerous research in neuroscience \citep{rayner1998eye,henderson1993eye}. In cognitively motivated NLP, several studies have investigated the impact of eye-tracking data on NLP tasks. In early works, these signals have been used in machine learning approaches to NLP tasks, such as part-of-speech tagging \citep{barrett2016weakly}, multiword expression extraction  \citep{DBLP:conf/ranlp/RohanianTYH17}, syntactic category prediction \citep{DBLP:conf/conll/BarrettS15}. In neural models, eye-tracking data are combined with word embeddings to improve various NLP tasks, such as sentiment analysis \citep{mishra2017leveraging} and NER \citep{DBLP:conf/naacl/HollensteinZ19}. Eye-tracking data have also been used to enhance or constrain neural attention in \citep{barrett2018sequence,sood2020improving,sood2020interpreting,DBLP:conf/emnlp/TakmazPBF20}. 

\paragraph{EEG for NLP.}
Electroencephalography (EEG) measures potentials fluctuations caused by the activity of neurons in cerebral cortex. The exploration of EEG data in NLP tasks is relatively limited. \citet{DBLP:conf/naacl/ChenCM12} improve the performance of automatic speech recognition (ASR) by using EEG signals to classify the speaker’s mental state. \citet{hollenstein2019advancing} incorporate EEG signals into NLP tasks, including NER, relation extraction and sentiment analysis. Additionally, \citet{DBLP:journals/corr/abs-2006-05113} leverage EEG features to regularize attention on relation extraction.
\paragraph{Adversarial Learning.}
The concept of adversarial training originates from the Generative Adversarial Nets (GAN) \citep{goodfellow2014generative} in computer vision. Since then, it has been also applied in NLP \citep{denton2015deep,ganin2016domain}. Recently, a great variety of studies attempt to introduce adversarial training into multi-task learning in NLP tasks, such as Chinese NER \citep{cao2018adversarial}, crowdsourcing learning \citep{DBLP:conf/aaai/YangZCZWZ18}, cross-lingual transfer learning \citep{chen2018adversarial,kim2017cross}, just name a few. Different from these studies, we use adversarial learning to deeply align cognitive modality to textual modality at the sentence level.

\section{CogAlign}
CogAlign is a general framework for incorporating cognitive processing signals into various NLP tasks. The target task can be specified at the predictor layer with corresponding task-specific neural network. CogAlign focuses on aligning cognitive processing signals to textual features at the word and encoder level. The text-aware attention aims at learning task-related useful cognitive information (thus filtering out noises) while the shared encoder and discriminator collectively learns to align representations of cognitive processing signals to those of textual inputs in a unified semantic space. The matched neural representations can be transferred to another datasets of the target task even though cognitive processing signals is not present. The neural architecture of CogAlign is visualized in Figure \ref{fig:basemodel}. We will elaborate the components of model in the following subsections. 
\subsection{Input Layer}
The inputs to our model include textual word embeddings and cognitive processing signals.
\paragraph{Word Embeddings.}
For a given word $x_i$ from the dataset of a target NLP task (e.g., NER), we obtain the vector representation $h^{word}_i$ by looking up a pre-trained embedding matrix. The obtained word embeddings are fixed during training. For NER, previous studies have shown that character-level features can improve the performance of sequence labeling \citep{DBLP:conf/acl/StoyanovJLY18}. We therefore apply a character-level CNN framework \citep{chiu2016named,DBLP:conf/acl/MaH16} to capture the character-level embedding. The word representation of word $x_i$ in NER task is the concatenation of word embedding and character-level embedding.
\paragraph{Cognitive Processing Signals.} For cognitive inputs, we can obtain word-level eye-tracking and EEG via data preprocessing (see details in Section 5.1). Thus, for each word $x_i$, we employ two cognitive processing signals $h^{eye}_i$ and $h^{eeg}_i$. The cognitive input $h^{cog}_i$ can be either a single type of signal or a concatenation of different cognitive processing signals.
\subsection{Text-Aware Attention}
As not all information contained in cognitive processing signals is useful for the target NLP task, we propose a text-aware attention mechanism to assign text sensitive weights to cognitive processing signals. The main process of attention mechanism consists of learning a compatibility matrix between word embeddings $H^{word}\in{R^{d_w\times{N}}}$ and cognitive representations $H^{cog}\in{R^{d_c\times{N}}}$ from the input layer and preforming cognitive-wise max-pooling operation over the matrix. The compatibility matrix $G\in{R^{d_w\times{d_c}}}$ can be computed as follows:
\begin{equation}\label{equ1}
G={\rm tanh}(H^{word}UH^{cog^{T}})
\end{equation}
where $d_w$ and $d_c$ are the dimension of word embeddings and cognitive representations, respectively, N is the length of the input, and $U\in{R^{N\times{N}}}$ is a trainable parameter matrix. 

We then obtain a vector $g^{cog}\in{R^{d_c}}$, which is computed as the importance score for each element in the cognitive processing signals with regard to the word embeddings, by row-wise max-pooling over $G$. Finally, we compute attention weights and the text-aware representation of cognitive processing signals ${H^{cog}}^{'}$ as follows:
\begin{equation}\label{equ7}
\alpha^{cog}={\rm softmax}(g^{cog})
\end{equation}
\begin{equation}\label{equ8}
{H^{cog}}^{'}=\alpha^{cog}H^{cog}
\end{equation}
\subsection{Encoder Layer}
We adopt Bi-LSTMs to encode both cognitive and textual inputs following previous works \citep{DBLP:conf/naacl/HollensteinZ19,hollenstein2019advancing}. In this work, we employ two private Bi-LSTMs and one shared Bi-LSTM as shown in Figure \ref{fig:basemodel}, where private Bi-LSTMs are used to encode cognitive and textual inputs respectively and the shared Bi-LSTM is used for learning shared semantics of both types of inputs. We concatenate the outputs of private Bi-LSTMs and shared Bi-LSTM as input to the task-specific predictors of subsequent NLP tasks. The hidden states of the shared Bi-LSTM are also fed into the discriminator.
\subsection{Modality Discriminator}
We alternatively feed cognitive and textual inputs into the shared Bi-LSTM encoder. Our goal is that the shared encoder is able to map the representations of the two different sources of inputs into the same semantic space so as to learn the deep commonalities of two modalities (cognitive and textual). For this, we use a self-supervised discriminator to provide supervision for training the shared encoder. 

Particularly, the discriminator is acted as a classifier to categorize the alternatively fed inputs into either the textual or cognitive input. For the hidden state of modality $k$, we use a self-attention mechanism to first reduce the dimension of the output of the shared Bi-LSTM $H^{s}_k\in{R^{d_h\times{N}}}$:

\begin{equation}\label{equ10}
\alpha={\rm softmax}(v^{T}{\rm tanh}(W_sH^{s}_k+b_s))
\end{equation}
\begin{equation}\label{equ11}
h^{s}_k=\sum^{N}_{i=1}\alpha_iH^{s}_{k_i}
\end{equation}
where $W_s\in{R^{d_h\times{d_h}}}$, $b_s\in{R^{d_h}}$, $v\in{R^{d_h}}$ are trainable parameters in the model, $h^{s}_k$ is the output of self-attention mechanism. Then we predict the category of the input by softmax function:
\begin{equation}\label{equ12}
D(h^{s}_k)={\rm softmax}(W_dh^{s}_k+b_d)
\end{equation}
where $D(h^{s}_k)$ is the probability that the shared encoder is encoding an input with modality $k$. 

\subsection{Predictor Layer}
Given a sample $X$, the final cognitively augmented representation after the encoder layer can be formulated as $H^{'}=[H^{p};H^{s}]\in{R^{2d_h\times{N}}}$. $H^{p}$ and $H^{s}$ are the result of private Bi-LSTM and shared Bi-LSTM, respectively. 

For sequence labeling tasks like NER, we employ the conditional random field (CRF) \citep{DBLP:conf/icml/LaffertyMP01} as the predictor as Bi-LSTM-CRF is widely used in many sequence labeling tasks \citep{DBLP:conf/acl/MaH16,DBLP:journals/bioinformatics/LuoYYZWLW18} due to the excellent performance and also in cognitively inspired NLP \citep{DBLP:conf/naacl/HollensteinZ19, hollenstein2019advancing}. Firstly, we project the feature representation $H^{'}$ onto another space of which dimension is equal to the number of NER tags as follows:
\begin{equation}\label{equ4}
o_i=W_nh^{'}_i+b_n
\end{equation}

We then compute the score of a predicted tag sequence $y$ for the given sample $X$:
\begin{equation}\label{equ5}
score(X,y)=\sum^{N}_{i=1}(o_{i,y_i}+T_{y_{i-1},y_i})
\end{equation}
where $T$ is a transition score matrix which defines the transition probability of two successive labels.

Sentiment analysis and relation extraction can be regarded as multi-class classification tasks, with 3 and 11 classes, respectively. For these two tasks, we use a self attention mechanism to  reduce the dimension of $H^{'}$ and obtain the probability of a predicted class via the softmax function.

\section{Training and Inference}
\subsection{Adversarial Learning}
In order to learn the deep interaction between cognitive and textual modalities in the same semantic space, we want the shared Bi-LSTM encoder to output representations that can fool the discriminator. Therefore we adopt the adversarial learning strategy. Particularly, the shared encoder acts as the generator that tries to align the textual and cognitive modalities as close as possible so as to mislead the discriminator. The shared encoder and discriminator works in an adversarial way.

Additionally, to further increase the difficulty for the discriminator to distinguish  modalities, we add a gradient reversal layer (GRL)  \citep{DBLP:conf/icml/GaninL15} in between the encoder layer and predictor layer. The gradient reversal layer does nothing in the forward pass but reverses the gradients and passes them to the preceding layer during the backward pass. That is, gradients with respect to the adversarial loss $\frac{\partial L_{Adv}}{\partial \theta}$ are replaced with $-\frac{\partial L_{Adv}}{\partial \theta}$ after going through GRL.

\subsection{Training Objective}
CogAlign is established on a multi-task learning framework, where the final training objective is composed of the adversarial loss $L_{Adv}$ and the loss of the target task $L_{Task}$. For NER, we exploit the negative log-likelihood objective as the loss function. Given $T$ training examples $(X^{i};\overline{y}^{i})$\footnote{$X$ can be either textual or cognitive input as we alternatively feed word embeddings and cognitive processing signals into CogAlign.}, $L_{Task}$ is defined as follows:
\begin{equation}\label{equ15}
L_{Task}=-\sum^{T}_{i=1}{\rm log}p(\overline{y}^{i}|X^{i})
\end{equation}
where $\overline{y}$ denotes the ground-truth tag sequence. The probability of $\overline{y}$ is computed by the softmax function:
\begin{equation}\label{equ16}
p(\overline{y}|X)=\frac{e^{score(X,\overline{y})}}{\sum_{\widetilde{y}\in{Y}}e^{score(X,\widetilde{y})}}
\end{equation}

For sentiment analysis and relation extraction tasks, the task objective is similar to that of NER. The only difference is that the label of the task is changed from a tag sequence to a single class.

The adversarial loss $L_{Adv}$ is defined as:
\begin{equation}\label{equ17}
L_{Adv}=\min_{\theta_s}(\max_{\theta_d}\sum^{K}_{k=1}\sum^{T_k}_{i=1}{\rm log}D(S(X^{i}_k)))
\end{equation}
where $\theta_s$ and $\theta_d$ denote the parameters of the shared Bi-LSTM encoders $S$ and modality discriminator $D$, respectively, $X^{i}_k$ is the representation of sentence $i$ in a modality $k$. The joint loss of CogAlign is therefore defined as:
\begin{equation}\label{equ18}
L=L_{Task}+L_{Adv}
\end{equation}
\subsection{Inference}
After training, the shared encoder learns a unified semantic space for representations of  both cognitive and textual modality. We believe that the shared space embeds knowledge from cognitive processing signals. For inference, we therefore only use the textual part and the shared encoder (components in the red dashed box in Figure \ref{fig:basemodel}). The private encoder outputs textual-modality-only representations while the shared encoder generates cognitive-augmented representations. The two representations are concatenated to feed into the predictor layer of the target task. This indicates that we do not need cognitive processing signals for the inference of the target task. It also means that we can pretrain CogAlign with cognitive processing signals and then transfer it to other datasets where cognitive processing signals are not available for the same target task.
\section{Experiments}
We conducted experiments on three NLP tasks, namely NER, sentiment analysis and relation extraction with two types of cognitive processing signals (eye-tracking and EEG) to validate the effectiveness of the proposed CogAlign. 

\begin{table*}[ht]
\small
\centering
\begin{tabular}{lll}
\hline
\multirow{2}{*}{\textbf{EARLY}}
& first fixation duration (FFD) & the duration of word $w$ that is first fixated\\
& first pass duration (FPD) & the sum of the fixations before eyes leave the word $w$\\
\hline
\multirow{6}{*}{\textbf{LATE}}
& number of fixations (NFIX) & the number of times word $w$ that is fixated \\
& fixation probability (FP) & the probability that word $w$ is fixated \\
& mean fixation duration (MFD) & the average fixation durations for word $w$\\
& total fixation duration (TFD) & the total duration of word $w$ that is fixated \\
& $n$ re-fixations (NR) & the number of times word $w$ that is fixated after the first fixation \\
& re-read probability (RRP) & the probability of word $w$ that is fixated more than once \\
\hline
\multirow{9}{*}{\textbf{CONTEXT}}
& total regression-from duration (TRD) & the total duration of regressions from word $w$ \\
& $w$-2 fixation probability ($w$-2 FP) & the fixation probability of the word $w$-2 \\
& $w$-1 fixation probability ($w$-1 FP) & the fixation probability of the word $w$-1 \\
& $w$+1 fixation probability ($w$+1 FP) & the fixation probability of the word $w$+1 \\
& $w$+2 fixation probability ($w$+2 FP) & the fixation probability of the word $w$+2 \\
& $w$-2 fixation duration ($w$-2 FD) & the fixation duration of the word $w$-2 \\
& $w$-1 fixation duration ($w$-1 FD) & the fixation duration of the word $w$-1 \\
& $w$+1 fixation duration ($w$+1 FD) & the fixation duration of the word $w$+1 \\
& $w$+2 fixation duration ($w$+2 FD) & the fixation duration of the word $w$+2 \\
\hline
\end{tabular}
\caption{Eye-tracking features used in the NER task.}\label{Tab:eye-tracking features}
\end{table*}
\subsection{Dataset and Cognitive Processing Signals}
We chose a dataset\footnote{The data is available here: https://osf.io/q3zws/} with multiple cognitive processing signals: Zurich Cognitive Language Processing Corpus (ZuCo) \citep{hollenstein2018zuco}. This corpus contains simultaneous eye-tracking and EEG signals collected when 12 native English speakers are reading 1,100 English sentences. Word-level signals can be divided by the duration of each word.

The dataset includes two reading paradigms: normal reading and task-specific reading where subjects exercise some specific task. In this work, we only used the data of normal reading, since this paradigm accords with human natural reading. The materials for normal reading paradigm consist of two datasets: 400 movie reviews from Stanford Sentiment Treebank \citep{socher2013recursive} with manually annotated sentiment labels, including 123 neutral, 137 negative and 140 positive sentences; 300 paragraphs about famous people from Wikipedia relation extraction corpus \citep{culotta2006integrating} labeled with 11 relationship types, such as award, education.

We also tested our model on NER task. For NER, the selected 700 sentences in the above two tasks are annotated with three types of entities: PERSON, ORGANIZATION, and LOCATION. All annotated datasets\footnote{https://github.com/DS3Lab/zuco-nlp/} are publicly available. The cognitive processing signals and textual features used for each task in this work are the same as \citep{hollenstein2019advancing}.

\paragraph{Eye-tracking Features.}
Eye-tracking signals record human gaze behavior while reading. The eye-tracking data of ZuCo are collected by an infrared video-based eye tracker EyeLink 1000 Plus with a sampling rate of 500 Hz. For NER, we used 17 eye-tracking features that cover all stages of gaze behaviors and the effect of context. According to the reading process, these features are divided into three groups:  \textbf{EARLY}, the gaze behavior when a word is fixated for the first time; \textbf{LATE}, the gaze behavior over a word that is fixated many times; \textbf{CONTEXT}, the eye-tracking features over neighboring words of the current word. The 17 eye-tracking features used in the NER task are shown in the Table \ref{Tab:eye-tracking features}.
In the other two tasks, we employed 5 gaze behaviors, including the first fixation duration (FFD), the number of fixations (NFIX), the total fixation duration (TFD), the first pass duration (FPD), the gaze duration (GD) that is the duration of the first time eyes move to the current word until eyes leave the word. 

\paragraph{EEG Features.}
EEG signals record the brain's electrical activity in the cerebral cortex by placing electrodes on the scalp of the subject. In the datasets we used, EEG signals are recorded by a 128-channel EEG Geodesic Hydrocel system (Electrical Geodesics, Eugene, Oregon) at a sampling rate of 500 Hz with a bandpass of 0.1 to 100 Hz. The original EEG signals recorded are of 128 dimensions. Among them, 23 EEG signals are removed during preprocessing since they are not related to the cognitive processing \citep{hollenstein2018zuco}. After preprocessing, we obtained 105 EEG signals. The left EEG signals are divided into 8 frequency bands by the frequency of brain's electrical signals: $theta1$ (t1, 4-6 Hz), $theta2$ (t2, 6.5-8 Hz), $alpha1$ (a1, 8.5-10 Hz), $alpha2$ (a2, 10.5-13 Hz), $beta1$ (b1, 13.5-18 Hz), $beta2$ (b2, 18.5-30 Hz), $gamma1$ (g1, 30.5-40 Hz) and $gamma2$ (g2, 40-49.5 Hz). The frequency bands reflects the different functions of brain cognitive processing. For NER, we used 8 EEG features that are obtained by averaging the 105 EEG signals at each frequency band. For the other two tasks, EEG features were obtained by averaging the 105 signals over all frequency bands. All used EEG features are obtained by averaging over all subjects and normalization. 
\begin{table*}[ht]
\scriptsize
\centering
\begin{tabular}{llccccccccc}
\hline
\multirow{2}{*}{\textbf{Signals}} & \multirow{2}{*}{\textbf{Model}} & \multicolumn{3}{c}{\textbf{NER}}&  \multicolumn{3}{c}{\textbf{Sentiment Analysis}} & \multicolumn{3}{c}{\textbf{Relation Extraction}}\\
\cline{3-5}   \cline{6-8} \cline{9-11}
~&~&P (\%)&R (\%)&F1 (\%)&P (\%)&R (\%)&F1 (\%)&P (\%)&R (\%)&F1 (\%)\\
~ & Base$^{*}$ & 89.34 & 78.60 & 83.48 & 59.47 & 59.42 & 58.27 & 79.52 & 75.67 & 75.25 \\
\hline
\multirow{4}{*}{eye}
& \citep{hollenstein2019advancing} & 86.2 & \textbf{84.3} & 85.1 & 61.4 & 61.7 & 61.5 & 65.1 & 61.9 & 62.0  \\
& Base & 90.56 & 81.05 & 85.43$^{*}$ & 64.26 & 61.96 & 61.19$^{*}$ & 82.01 & 78.23 & 77.95$^{*}$ \\
& Base+TA & 90.75 & 81.77 & 85.93$^{*}$ & 64.63 & 62.71 & 61.41$^{*}$ & \textbf{83.26} & 76.47 & 78.04$^{*}$\\
& CogAlign & 90.76 & 82.52 & 86.41$^{*}$ & 62.86 & 64.10 & 62.30$^{*}$ & 78.33 & 82.06 & 78.56$^{*}$ \\
\hline
\multirow{4}{*}{EEG}
& \citep{hollenstein2019advancing} & 86.7 & 81.5 & 83.9 & 60.5 & 60.2 & 60.3 & 68.3 & 64.8 & 65.1\\
& Base & 89.82 & 80.55 & 84.76$^{*}$ & 64.09 & 60.29 & 59.79$^{*}$ & 82.79 & 77.16 & 77.61$^{*}$ \\
& Base+TA & 89.54 & 82.22 & 85.62$^{*}$ & 62.20 & 62.19 & 60.91$^{*}$ & 80.83 & 78.46 & 77.81$^{*}$\\
& CogAlign & 89.87 & 83.08 & 86.21$^{*}$ & 63.11 & 65.38 & 62.81$^{*}$ & 77.94 & \textbf{82.60} & 78.66$^{*}$\\
\hline
\multirow{4}{*}{\shortstack{eye\\+EEG}}
& \citep{hollenstein2019advancing} & 85.1 & 83.2 & 84.0 & 59.8 & 60.0 & 59.8 & 66.3 & 59.3 & 60.8\\
& Base & 89.70 & 81.11 & 85.11$^{*}$ & 62.86 & 61.49 & 60.84$^{*}$ & 79.00 & 76.52 & 77.72$^{*}$\\
& Base+TA & 90.75 & 82.94 & 86.31$^{*}$ & \textbf{65.22} & 63.88 & 63.23$^{*}$ & 82.24 & 77.53 & 78.12$^{*}$\\
& CogAlign & \textbf{91.28} & 83.02 & \textbf{86.79}$^{*}$ & 65.11 & \textbf{65.94} & \textbf{65.40}$^{*}$ & 78.66 & 82.07 & \textbf{78.93}$^{*}$\\
\hline
\end{tabular}
\caption{Results of CogAlign and other methods on the three NLP tasks augmented with eye-tracking features (eye), EEG features (EEG), and both (eye+EEG). ‘Base$^{*}$' denotes that the model does not use any cognitive processing signals. ‘Base' is a neural model that consist of a textual private encoder and textual predictor, and combines cognitive processing signals with word embeddings via direct concatenation, similar to previous works. ‘Base+TA' is a neural model where direct concatenation in the base model is replaced by the text-aware attention mechanism. Significance is indicated with the asterisks: * = p\textless0.01.}\label{Tab:1}
\end{table*}
\subsection{Settings}
We evaluated three NLP tasks in terms of precision, recall and F1 in our experiments. Word embeddings of all NLP tasks were initialized with the publicly available pretrained GloVe \citep{DBLP:conf/emnlp/PenningtonSM14} vectors of 300 dimensions. For NER, we used 30-dimensional randomly initialized character embeddings. We set the dimension of hidden states of LSTM to 50 for both the private Bi-LSTM and shared Bi-LSTM. We performed 10-fold cross validation for NER and sentiment analysis and 5-fold cross validation for relation extraction.

\subsection{Baselines}
We compared our model with previous state-of-the-art methods on ZuCo dataset. The method by \citet{hollenstein2019advancing} incorporates cognitive processing signals into their model via direct concatenation mentioned before.

\subsection{Results}
Results of CogAlign on the three NLP tasks are shown in Table \ref{Tab:1}. From the table, we observe that:
\begin{itemize}
\item By just simply concatenating word embeddings with cognitive processing signals, the Base model is better than the model without using any cognitive processing signals, indicating that cognitive processing signals (either eye-tracking or EEG signals) can improve all three NLP tasks. Notably, the improvements gained by eye-tracking features are larger than those obtained by EEG signals while the combination of both does not improve over only using one of them. We conjecture that this may be due to the low signal-to-noise ratio of EEG signals, which further decreases when two signals are combined together.

\item Compared with the Base model, the Base+TA achieves better results on all NLP tasks. The text-aware attention gains an absolute improvement of 0.88, 2.04, 0.17 F1 on NER, sentiment analysis, and relation extraction, respectively. With Base+TA, the best results for most tasks are obtained by the combination of eye-tracking and EEG signals. This suggests that the proposed text-aware attention may have alleviated the noise problem of cognitive processing signals.
\item The proposed CogAlign achieves the highest F1 over all three tasks, with  improvements of 0.48, 2.17 and 0.87 F1 over Base+TA on NER, sentiment analysis and relation extraction, respectively, which demonstrates the effectiveness of our proposed model. In addition, CogAlign with both cognitive processing signals obtains new state-of-the-art performance in all NLP tasks. This suggests that CogAlign is able to effectively augment neural models with cognitive processing signals.
\end{itemize}

\begin{table*}[ht]
\small
\centering
\begin{tabular}{lccccccccc}
\hline
\multirow{2}{*}{\textbf{Model}} &
\multicolumn{3}{c}{\textbf{NER}}&  \multicolumn{3}{c}{\textbf{Sentiment Analysis}} & \multicolumn{3}{c}{\textbf{Relation Extraction}}\\
\cline{2-4}   \cline{5-7} \cline{8-10}
~&P (\%)&R (\%)&F1 (\%)&P (\%)&R (\%)&F1 (\%)&P (\%)&R (\%)&F1 (\%)\\
\hline
CogAlign (eye+EEG) & 91.28 & 83.02 & 86.79$^{*}$ & 65.11 & 65.94 & 65.40$^{*}$ & 78.66 & 82.07 & 78.93$^{*}$\\
- text-aware attention & 90.51 & 82.45 & 86.19$^{*}$ & 64.75 & 65.30 & 63.90$^{*}$ & 77.67 & 83.14 & 78.68$^{*}$\\
- cognitive loss & 90.20 & 81.11 & 85.45$^{*}$ & 64.48 & 65.42 & 63.77$^{*}$ & 77.79 & 81.24 & 77.75$^{*}$\\
- modality discriminator & 89.63 & 83.66 & 86.09$^{*}$ & 64.11 & 66.24 & 63.28$^{*}$ & 78.61 & 80.71 & 78.46$^{*}$\\
\hline
\end{tabular}
\caption{Ablation study on the three NLP tasks. Significance is indicated with the asterisks: * = p\textless0.01.}\label{Tab:2}
\end{table*}

\subsection{Ablation Study}

To take a deep look into the improvements contributed by each part of our model, we perform ablation study on all three NLP tasks with two cognitive processing signals. The ablation test includes: (1) \textbf{w/o text-aware attention}, removing text-aware attention mechanism; (2) \textbf{w/o cognitive loss}, discarding the loss of the cognitive predictor whose inputs are cognitive processing signals; (3) \textbf{w/o modality discriminator}, removing the discriminator to train parameters with the task loss. Table \ref{Tab:2} reports the ablation study results.

\begin{figure}[ht]
\centering
\subfigure[without adv]
{
	\includegraphics[width=0.2\textwidth]{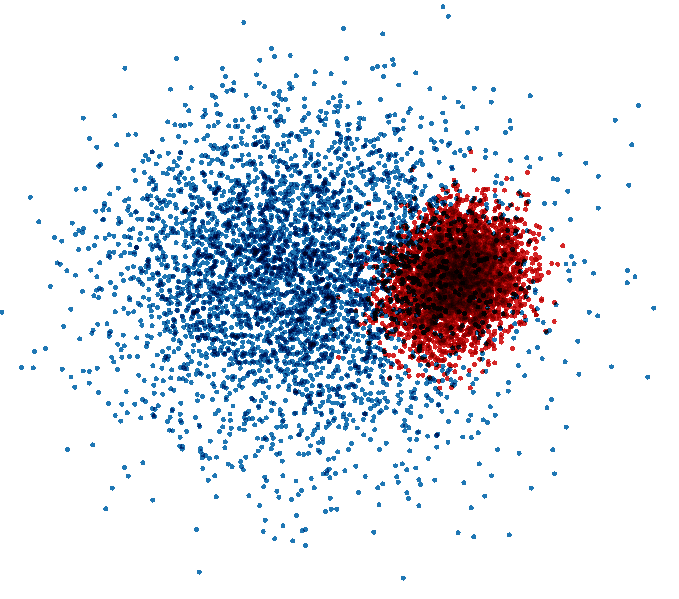}
}
\subfigure[with adv]
{
	\includegraphics[width=0.2\textwidth]{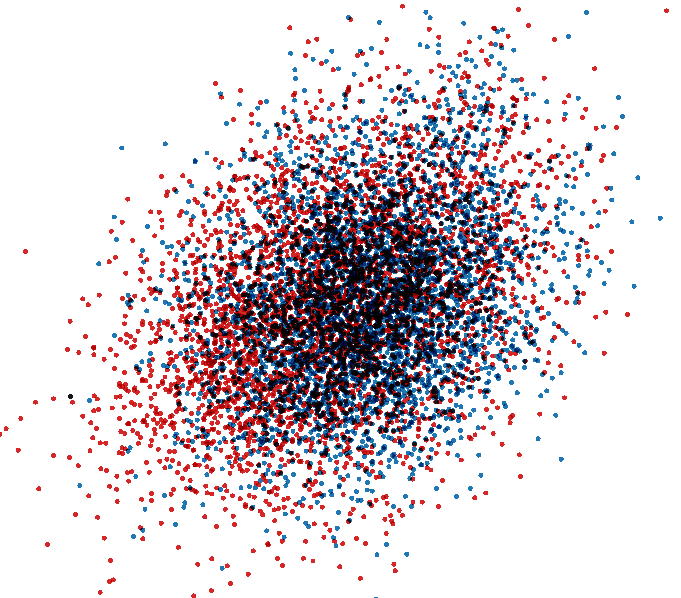}
}
\caption{The visualization of hidden states from the shared Bi-LSTM layer. ‘adv' denotes the adversarial learning. Red dots are the hidden representations of cognitive processing signals while blue dots hidden representations of textual inputs. Both are at the word level via t-SNE \citep{van2008visualizing}.}\label{fig:2}
\end{figure}

The absence of the text-aware attention, cognitive loss and modality discriminator results in a significant drop in performance. This demonstrates that these components all contribute to the effective incorporation of cognitive processing signals into neural models of the three target tasks. CogAlign outperforms both
(2) \textbf{w/o cognitive loss} and (3) \textbf{w/o modality discriminator}
by a great margin, indicating that the cognitive features can significantly enhance neural models. 

Furthermore, we visualize the distribution of hidden states learned by the shared Bi-LSTM to give a more intuitive demonstration of the effect of adversarial learning. In Figure \ref{fig:2}, clearly, the modality discriminator with adversarial learning forces the shared Bi-LSTM encoder to align textual inputs to cognitive processing signals in the same space. 

\section{Analysis}
\subsection{Text-aware Attention Analysis}
\begin{figure}[ht]
\centering
\subfigure[eye-tracking]
{
	\includegraphics[width=0.4\textwidth]{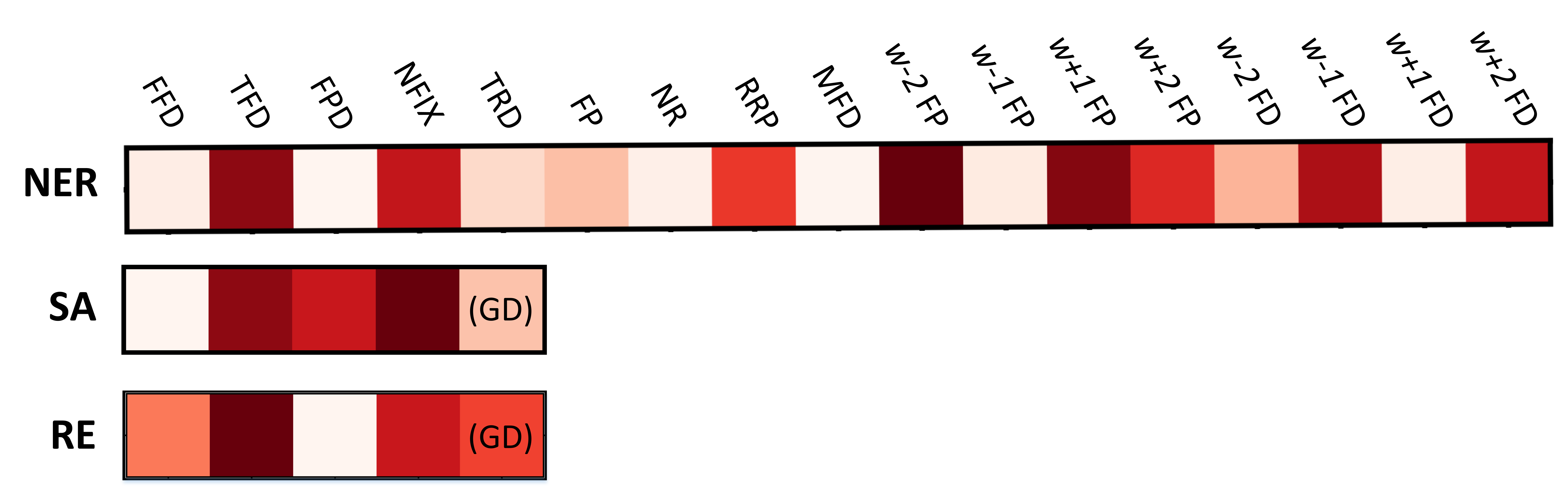}\label{fig:3(a)}
}
\subfigure[EEG]
{
	\includegraphics[width=0.25\textwidth]{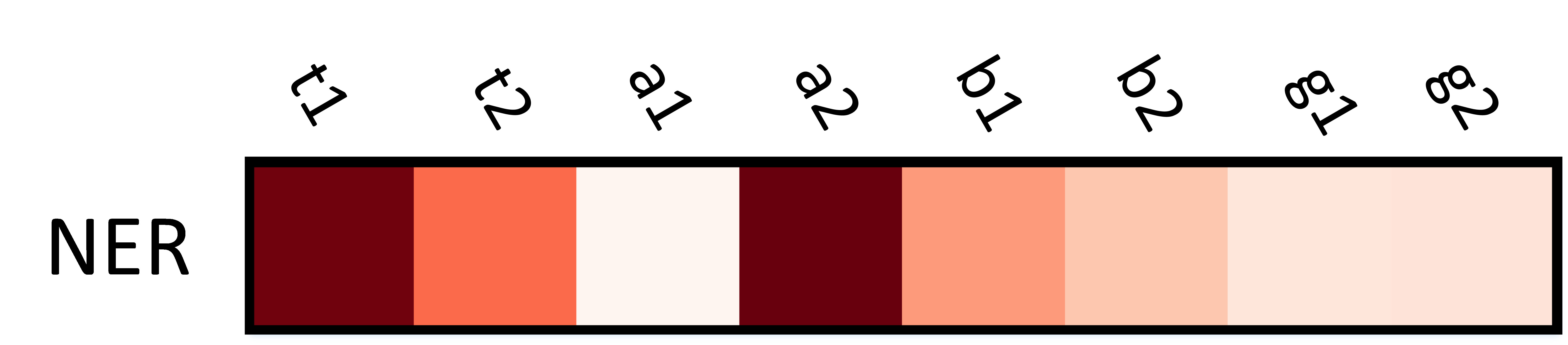}\label{fig:3(b)}
}
\caption{The visualization of attention weights over cognitive processing signals by the text-aware attention in the three NLP tasks. Darker colors represent higher attention weights.}\label{fig:3}
\end{figure}

In addition to denoising the cognitive processing signals, the text-aware attention mechanism also obtains the task-specific features. To have a clear view of the role that the text-aware attention mechanism plays in CogAlign, we randomly choose samples and visualize the average attention weights over each signal in Figure \ref{fig:3}. 

For eye-tracking, signals reflecting the late syntactic processing, such as ‘NFIX' (number of fixation), ‘TFD' (total fixation duration), play an important role in the three tasks. These results are consistent with findings in cognitive neuroscience. In cognitive neuroscience, researchers have shown that readers tend to gaze at nouns repeatedly \citep{2009Nomen} (related to the eye-tracking signal NFIX, the number of fixations) and there is a dependency relationship between regression features and sentence syntactic structures \citep{2019Dependency}. In other NLP tasks that infused eye-tracking features, the late gaze features have also proved to be more important than early gaze features, such as multiword expression extraction \citep{DBLP:conf/ranlp/RohanianTYH17}. Moreover, from the additional eye-tracking used in NER, we can find that the cognitive features from the neighboring words are helpful to identify entity, such as ‘$w$-2 FP' ($w$-2 fixation probability), ‘$w$+1 FP' ($w$+1 fixation probability).

Since a single EEG signal has no practical meaning, we only visualize the attention weights over EEG signals used in the NER task. Obviously, attentions to ‘t1' ($theta1$) and ‘a2' ($alpha2$) are stronger than other signals, suggesting that low frequency electric activities in the brain are obvious when we recognize an entity.

\begin{table*}[ht]
\small
\centering
\begin{tabular}{lcccccc}
\hline
\multirow{2}{*}{\textbf{Model}} &
\multicolumn{3}{c}{\textbf{Wikigold}}&  \multicolumn{3}{c}{\textbf{SST}}\\
\cline{2-4}   \cline{5-7}
~&P (\%)&R (\%)&F1 (\%)&P (\%)&R (\%)&F1 (\%)\\
\hline
baseline & 80.70 & 70.67 & 75.19 & 56.67 & 57.58 & 56.40\\
baseline (+ZuCo text) &77.95 & 76.07 & 76.42 & 58.13 & 58.04 & 56.99 \\
baseline (two encoders) & 80.16 & 73.39 & 75.73 & 56.76 & 58.05 & 56.89 \\
CogAlign (eye) & 80.39 & 72.59 & 76.17 & 58.05 & 59.69 & 57.27 \\
CogAlign (EEG) & 80.54 & 71.91 & 75.93 & 57.25 & 58.34 & 57.10 \\
CogAlign (eye+EEG) & 81.71 & 74.17 & 77.76 & 58.60 & 58.33 & 58.32 \\
\hline
\end{tabular}
\caption{Results of CogAlign in transfer learning to other datasets  without cognitive processing signals. ‘baseline' is a model trained and tested with one encoder for textual inputs. ‘baseline (+ZuCo text)’ is the baseline trained with both Zuco textual data and target dataset (i.e., Wikigold or SST). ‘baseline (two encoders)’ is the same as CogAlign (the inference version), where cognitive processing signals are replaced by textual inputs.}\label{Tab:3}
\end{table*}
\subsection{Transfer Learning Analysis}
The cognitively-inspired NLP is limited by the collection of cognitive processing signals. Thus, we further investigate whether our model can transfer cognitive features to other datasets without cognitive processing signals for the same task. We enable transfer learning in CogAlign with a method similar to the alternating training approach \citep{DBLP:journals/corr/LuongLSVK15} that optimizes each task for a fixed number of mini-batches before shifting to the next task. In our case, we alternately feed instances from the ZuCo dataset and those from other datasets built for the same target task but without cognitive processing signals into CogAlign. Since CogAlign is a multi-task learning framework, model parameters can be updated either by data with cognitive processing signals or by data without such signals, where task-specific loss is used in both situations. Please notice that only textual inputs are fed into trained CogAlign for inference. 

To evaluate the capacity of CogAlign in transferring cognitive features, we select benchmark datasets for NER and sentiment analysis: Wikigold \citep{DBLP:conf/acl-pwnlp/BalasuriyaRNMC09} and Stanford Sentiment Treebank \citep{socher2013recursive}. Since no other datasets use the same set of relation types as that in ZuCo dataset, we do not test the relation extraction task for transfer learning. To ensure that the same textual data are used for comparison, we add a new baseline model (baseline (+Zuco text)) that is trained on the combination of textual data in ZuCo and benchmark dataset. Additionally, as CogAlign uses two encoders for inference (i.e., the textual encoder and shared encoder), for a fair comparison, we setup another baseline (baseline (two encoders)) that also uses two encoders fed with the same textual inputs. The experimental setup is the same as mentioned before.

Results are shown in the Table \ref{Tab:3}. We can observe that CogAlign consistently outperforms the two baselines. It indicates that CogAlign is able to effectively transfer cognitive knowledge (either eye-tracking or EEG) from ZuCo to other datasets. Results show that the best performance is achieved by transferring both eye-tracking and EEG signals at the same time. 

\section{Conclusions}
In this paper, we have presented CogAlign, a framework that can effectively fuse cognitive processing signals into neural models of various NLP tasks by learning to align the textual and cognitive modality at both word and sentence level. Experiments demonstrate that CogAlign achieves new state-of-the-art results on three NLP tasks on the Zuco dataset. Analyses suggest that the text-aware attention in CogAlign can learn task-related cognitive processing signals by attention weights while the modality discriminator with adversarial learning forces CogAlign to learn cognitive and textual representations in the unified space. Further experiments exhibit that CogAlign is able to transfer cognitive information from Zuco to other datasets without cognitive processing signals.

\section*{Acknowledgments}
The present research was partially supported by the National Key Research and Development Program of China (Grant No. 2019QY1802) and Natural Science Foundation of Tianjin (Grant No. 19JCZDJC31400). We would like to thank the anonymous reviewers for their insightful comments. 
\bibliographystyle{acl_natbib}
\bibliography{anthology,acl2021}


\end{document}